\documentclass{article}

\usepackage{arxiv}

\usepackage[utf8]{inputenc} % allow utf-8 input
\usepackage[T1]{fontenc}    % use 8-bit T1 fonts
\usepackage{hyperref}       % hyperlinks
\usepackage{url}            % simple URL typesetting
\usepackage{booktabs}       % professional-quality tables
\usepackage{amsfonts}       % blackboard math symbols
\usepackage{nicefrac}       % compact symbols for 1/2, etc.
\usepackage{microtype}      % microtypography
\usepackage{lipsum}
\usepackage{graphicx}
\usepackage{amsmath}
\usepackage{float} % 允许使用 H 选项
\usepackage{multirow}
\graphicspath{ {./images/} }

\title{Generative Reliability-Based Design Optimization Using In-Context Learning Capabilities of Large Language Models}

\author{
 Zhonglin Jiang \\
  School of Mechanical and Electrical Engineering\\
  University of Electronic Science and Technology of China\\
  2006 Xiyuan Ave, West Zone, High-Tech Zone, Chengdu, China \\
  \texttt{202322040554@std.uestc.edu.cn} \\
   \And
 Qian Tang \\
  School of Mechanical and Electrical Engineering\\
  University of Electronic Science and Technology of China\\
  2006 Xiyuan Ave, West Zone, High-Tech Zone, Chengdu, China \\
  \texttt{202322040552@std.uestc.edu.cn} \\
   \And
 Zequn Wang \\
  School of Mechanical and Electrical Engineering\\
  University of Electronic Science and Technology of China\\
  2006 Xiyuan Ave, West Zone, High-Tech Zone, Chengdu, China \\
  \texttt{zequnwang@uestc.edu.cn} \\
}

\begin{document}
\maketitle
\begin{abstract}
Large Language Models (LLMs) have demonstrated remarkable in-context learning capabilities, enabling flexible utilization of limited historical information to play pivotal roles in reasoning, problem-solving, and complex pattern recognition tasks. Inspired by the successful applications of LLMs in multiple domains, this paper proposes a generative design method by leveraging the in-context learning capabilities of LLMs with the iterative search mechanisms of metaheuristic algorithms for solving reliability-based design optimization problems. In detail, reliability analysis is performed by engaging the LLMs and Kriging surrogate modeling to overcome the computational burden. By dynamically providing critical information of design points to the LLMs with prompt engineering, the method enables rapid generation of high-quality design alternatives that satisfy reliability constraints while achieving performance optimization. With the Deepseek-V3 model, three case studies are used to demonstrated the performance of the proposed approach. Experimental results indicate that the proposed LLM-RBDO method successfully identifies feasible solutions that meet reliability constraints while achieving a comparable convergence rate compared to traditional genetic algorithms. 
\end{abstract}

% keywords can be removed

\section{Introduction}
Mechanical design optimization constitutes a pivotal task in modern engineering, aiming to enhance product performance, efficiency, and cost-effectiveness through parametric optimization \cite{maalawi2009design}. With advancements in computational technology and optimization algorithms, design optimization has become standard practice in mechanical engineering. However, traditional deterministic optimization methods often neglect external uncertainties such as material properties, geometric variations, assembly conditions, and operational loads \cite{wang2016validating}. Reliability-Based Design Optimization addresses this limitation by incorporating probabilistic constraints, ensuring that designs maintain specified reliability levels under various uncertainty conditions throughout the optimization process \cite{ling2022overview}.

The traditional RBDO framework consists of three key components: performance function evaluation, reliability analysis, and design optimization \cite{hu2024reliability}. Within RBDO processes, performance function evaluation serves dual purposes: quantifying current design performance and elucidating uncertainty propagation mechanisms. This facilitates precise sensitivity analysis of failure probabilities relative to design variables, thereby enhancing optimization decision-making \cite{del2019cutting}. Three predominant evaluation approaches exist: Direct computation via simulations or physical formulations \cite{li2019extending}; Data-driven modeling techniques including Kriging models and artificial neural networks \cite{wang2014maximum}; Hybrid models combining physical principles with data-driven methods to improve evaluation accuracy \cite{xu2022multi}. In RBDO, reliability analysis plays a critical role in identifying potential failure modes, estimating system lifespan, and assessing reliability levels \cite{zhang2020hybrid}. For time-invariant uncertainty scenarios, reliability is typically analyzed through failure probability computation using methods such as First-Order Reliability Method (FORM) \cite{chen2025novel}, Second-Order Reliability Method (SORM) \cite{lou2025reliability}, and Monte Carlo Simulation (MCS) \cite{breve2023flow}. Time-dependent uncertainty analysis employs specialized techniques including outcrossing rate method \cite{zhang2024improved} and extreme value method \cite{song2022estimation}. Reliability-based design optimization strategies primarily encompass the double-loop approach, decoupling methods, and single-loop techniques \cite{zhang2021reliability,keshtegar2018enhanced,shi2020novel}. The double-loop approach represents the most fundamental method for addressing RBDO problems, featuring a tightly coupled double-nested structure where outer design variable optimization interacts iteratively with inner reliability analysis. However, its prohibitive computational cost and low efficiency have motivated the development of single-loop and decoupling alternatives. Single-loop methods circumvent nested reliability analysis cycles by substituting reliability constraints with approximate equivalence conditions. Decoupling methods enhance computational efficiency through sequential treatment of design variable optimization and reliability analysis, effectively disentangling these processes algorithmically. Building upon these foundational strategies, contemporary design optimization algorithms can be further categorized into gradient-based methods and meta-heuristic algorithms. Gradient-based algorithms identify optimal solutions by following the gradient directions of reliability constraints, with prominent implementations including classical gradient descent, Newton's method, Sequential Quadratic Programming (SQP), and Generalized Reduced Gradient (GRG) algorithms \cite{carlon2022risk,liu2023periodic,okoro2023dependency,shivashankar2022estimation}. However, these methods require performance functions and constraints to be continuously differentiable, while remaining susceptible to noise interference and local optimum entrapment \cite{hamza2022new,lopez2017reliability}. In contrast, metaheuristic algorithms demonstrate strong adaptability for handling discrete, combinatorial, and hybrid optimization problems, coupled with superior global search capabilities that have enabled their widespread adoption in reliability design optimization \cite{abualigah2022meta}. As a representative, Shahraki A. F. et al. \cite{shahraki2014reliability} proposed a general approach that combines genetic algorithms with a performance-measure-based reliability evaluation loop to generate reliable and robust Pareto-optimal solutions. S. Winyangkul et al. \cite{winyangkul2021reliability} introduced a fuzzy metaheuristic algorithm to address the RBDO problem of aircraft wings. J. Jafari-Asl et al. \cite{jafari2024meta} applied a reliability-based design optimization approach to shell-and-tube heat exchangers using a control-variable surrogate model and a hybrid metaheuristic algorithm. However, traditional metaheuristic algorithms, such as genetic algorithms, ant colony optimization, and particle swarm optimization, often suffer from high computational costs and potentially slow convergence rates \cite{almufti2023overview} .

Generative Artificial Intelligence (Generative AI) refers to a technology that generates new content by leveraging existing data. Unlike traditional discriminative models, whose goal is to classify or predict based on input features, generative models aim to learn the distribution of data and generate samples that resemble the training data \cite{feuerriegel2024generative}. The core of such models lies in their ability to understand the underlying structure of data and, based on this structure, generate new, innovative, and diverse samples. Notable techniques in generative AI include Generative Adversarial Networks (GANs) \cite{yuan2023dde}, Variational Autoencoders (VAEs) \cite{li2022predictive}, Diffusion Models \cite{shu2025df}, and Transformers \cite{li2024transformer}. The applications of generative AI are widespread. Kazemi et al. \cite{kazemi2022multiphysics} demonstrated that GANs could effectively predict topological structures similar to those generated by level set topology optimization (LSTO) by learning from optimal designs of single-physics problems and applying this knowledge to multi-physics problem coupling. Gladstone et al. \cite{gladstone2024robust} proposed a robust topology optimization method based on VAEs, showing its effectiveness in optimizing L-shaped bracket and heat sink designs under uncertain loading conditions. Li et al. \cite{li2024diffusion} utilized diffusion models to transform design optimization into a conditional sampling problem, introducing a reward-driven diffusion model to efficiently generate near-optimal design solutions while preserving the underlying structure.

Recently, Large Language Models represent a significant breakthrough in the field of generative AI. By leveraging massive parameter scales and pre-training on vast amounts of text data, LLMs exhibit unprecedented language understanding and generation capabilities \cite{zhao2023survey}. Unlike traditional language models, the core of LLMs lies in the self-attention mechanism based on the Transformer architecture \cite{vaswani2017attention}, which captures deep contextual relationships in language. This enables them to efficiently process long texts and comprehend complex semantics, a capability referred to as In-Context Learning (ICL) \cite{dong2022survey}. This ability allows LLMs to rapidly adapt to specific tasks with minimal examples or task descriptions, without the need for fine-tuning or large annotated datasets. As a result, LLMs can efficiently perform tasks such as text generation, problem-solving, and complex reasoning. Chen Y et al. \cite{chen2021meta} proposed a meta-learning method based on In-Context Tuning (ICT) using large language models, which transforms task adaptation and prediction problems into sequence prediction tasks. Liu S et al. \cite{liu2024large} explored the use of large language models as evolutionary combinatorial optimizers, studying their role in the classical Traveling Salesman Problem (TSP). Yin Y et al. \cite{yin2024ado} proposed a design optimization method combining large language models and Bayesian Optimization (BO), and demonstrated significant improvements in design efficiency and effectiveness through evaluations on two simulated circuits. Ferber D et al. \cite{ferber2024context} utilized the context learning capabilities of large language models to perform medical image classification tasks in organizational pathology. Zhang M R et al. \cite{zhang2023using} investigated the use of foundation large language models for hyperparameter optimization (HPO). Their empirical evaluations showed that under limited search budgets, LLMs could match or even outperform traditional methods such as random search and Bayesian optimization. AhmadiTeshnizi A et al. \cite{ahmaditeshnizi2024optimus} proposed OptiMUS-0.3, an LLM-based system capable of automatically solving linear programming problems from natural language descriptions. Experimental results indicated that the system achieved performance improvements of over $22 \%$ and $24 \%$ on simple and complex datasets, respectively. Liu F et al. \cite{liu2025large} leveraged carefully designed prompt engineering to enable general LLMs to serve as zero-shot black-box search operators for MOEA/D. They further analyzed LLM behavior to construct an explicit white-box operator, which demonstrated strong generalization ability on unseen problems. Guo Q et al. \cite{guo2024connectinglargelanguagemodels} introduced EvoPrompt, a discrete prompt optimization framework that integrates evolutionary algorithms (EAs) with LLMs. By iteratively generating new prompts and optimizing human-readable natural language expressions, this framework showcased the potential and synergy of combining LLMs with traditional algorithms.

Despite significant advancements in various fields, the application of large language models in engineering design remains relatively limited. Zhang X et al. \cite{zhang2024using} explored the potential of LLMs in parametric shape optimization (PSO), demonstrating faster convergence compared to traditional classical optimization algorithms. Lu J et al. \cite{lu2024constructing} proposed a method that guides LLMs to learn how to construct comprehensive mechanical design agents (MDAs), addressing issues such as high learning costs and repetitive tasks in current mechanical design. Jadhav Y et al. \cite{jadhav2024large} introduced an innovative approach that combines pre-trained LLMs with finite element method (FEM) modules to optimize the efficiency and applicability of traditional mechanical design. These studies provide strong support for the application of LLMs in the field of mechanical design. However, in the specific area of mechanical design optimization, particularly in handling complex reliability analysis and design optimization problems, the potential of LLMs remains to be further explored. Inspired by the success of large language models as decision-makers in various applications, we explore their potential in Reliability-Based Design Optimization. This paper presents a generative optimization method based on the context learning ability of large language models, combining the model's generative capability with the iterative search mechanism of metaheuristic algorithms, applied to reliability design optimization problems, referred to as LLM-RBDO. Through this approach, the model can rapidly generate high-quality design points, explore optimal solutions, and effectively address the challenges posed by uncertainty propagation and the complexity of objective functions, offering new solutions for optimization problems in engineering design.

\section{Related Works}
\begin{figure}
    \centering
    \includegraphics[width=0.5\linewidth]{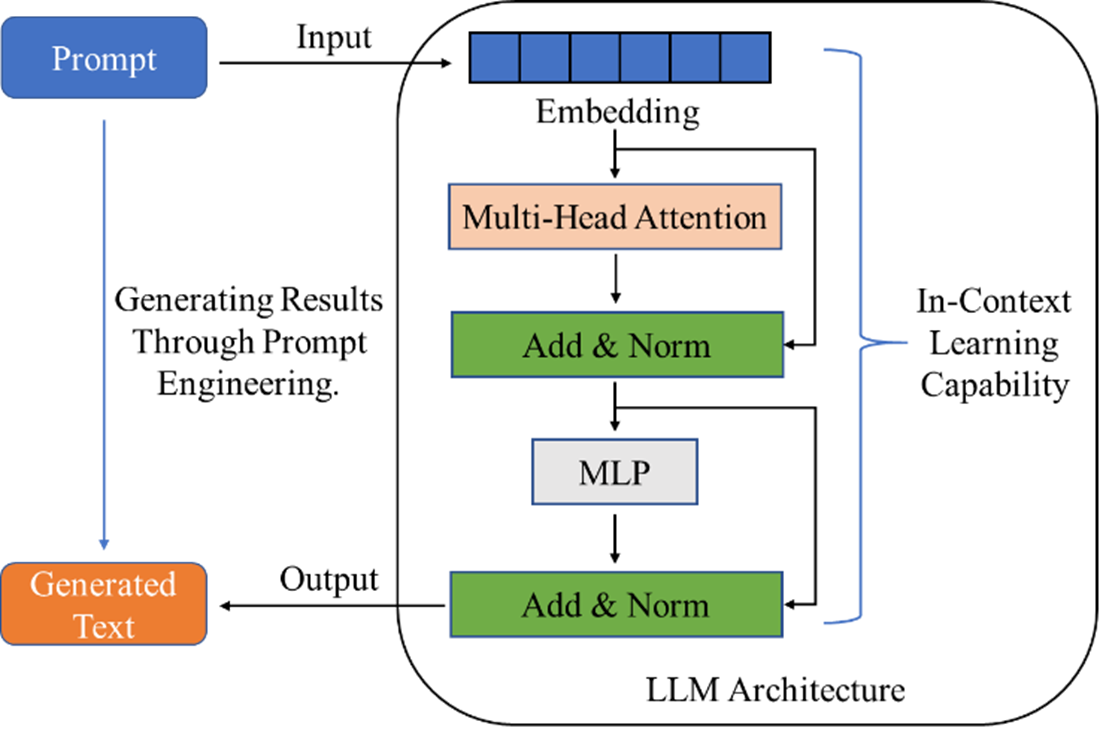}
    \caption{LLM Architecture and Prompt Engineering}
    \label{fig:1}
\end{figure}

This section primarily discusses the fundamental principles of LLM implementation and how prompts interact with LLMs through prompt engineering to generate text. The specific implementation is illustrated in Figure \ref{fig:1}, where prompts are encoded through embedding and fed into a multi-head attention mechanism. After further processing through additional layers, the final output is generated. This architecture endows large language models with the capability of contextual learning. There are various ways in which prompts interact with LLMs, and prompt engineering refers to the techniques used to process prompts and facilitate their interaction with LLMs. In detail, Subsection \ref{2.1} provides a detailed explanation of the origin and fundamental principles of LLMs, Subsection \ref{2.2} explores contextual learning and its application scenarios, and Subsection \ref{2.3} introduces the definition and functionality of prompt engineering.
\subsection{Background of LLMs}
\label{2.1}
Natural Language Processing (NLP) has evolved from rule-based approaches to data-driven methodologies. Early statistical models, such as n-gram models, relied on word frequency statistics for text prediction but were constrained by local dependencies and the curse of dimensionality. With the advancement of deep learning, Recurrent Neural Networks (RNNs) and their variants, including Long Short-Term Memory (LSTM) and Gated Recurrent Units (GRU), introduced sequential modeling to enable contextual awareness. However, these models still faced challenges in capturing long-range dependencies and were limited by inefficient parallelization during training. In 2017, Vaswani et al. introduced the foundational architecture for large language models: the Transformer. This architecture consists of two main components: an encoder and a decoder. The encoder extracts contextual information from the input sequence using a self-attention mechanism and transforms it into context-aware representations. The decoder then utilizes these representations to generate the output sequence. Within both the encoder and decoder, each layer comprises a multi-head attention mechanism and a feed-forward network. Attention is the core component of the Transformer, enabling it to capture dependencies across long sequences effectively. The attention mechanism operates using three key matrices: a query matrix $Q \in \mathbb{R}^{n \times d_k}$, a key matrix $K \in \mathbb{R}^{m \times d_k}$, and a value matrix$V \in \mathbb{R}^{m \times d_v}$, where each row of these matrices corresponds to a word in the sequence. The attention computation is performed as follows:
\begin{equation}
H=\operatorname{Attention}(Q, K, V)=\operatorname{Softmax}\left(\frac{Q K^T}{\sqrt{d_k}}\right) V
\end{equation}
where the three matrices $Q$, $K$, and $V$ are obtained through linear transformations: $Q=X_q W_q$, $V=X_{k v} W_k$, $V=X_{k v} W_v$. Here,$X_q \in \mathbb{R}^{n \times d}$ and$X_kv \in \mathbb{R}^{m \times d}$ represent the feature matrices of the query sequence (with sequence length n) and the key-value sequence (with sequence length m), respectively. The weight matrices $W_q$, $W_k$, and $W_v$ map the input feature space to the corresponding query, key, and value representations.

Moreover, the multi-head attention mechanism enables the model to capture diverse contextual information across different subspaces, thereby enhancing its representational capacity. With advances in computational power and the accumulation of large-scale corpora, models such as GPT and BERT have refined the attention masking mechanism—employing causal masking for autoregressive modeling and fully visible masking for autoencoding paradigms. As model parameter sizes surpassed the hundred-billion scale (e.g., GPT-3), they exhibited emergent abilities, including in-context learning, marking the advent of the large language model era.
\subsection{Incontext Learning}
\label{2.2}
In-Context Learning emerged as a novel learning paradigm alongside the development of GPT-3. Unlike traditional training methods, ICL enables models to learn task patterns without updating their parameters. Instead, the model infers patterns by processing a sequence of demonstrations within the input context and applying them to new queries. Specifically, ICL leverages a few-shot learning approach, where a small number of examples and test samples are provided in the input. By utilizing the pretrained knowledge and pattern recognition capabilities of LLMs, the model can directly generate responses that align with the demonstrated patterns. The core idea of ICL is to learn through analogy, where the model can infer implicit task rules from provided examples and perform reasoning without explicit training. This capability makes ICL a crucial subclass of Prompt Learning, where the examples form part of the prompt that guides the model to generate the desired output. By using these examples, the model can generalize patterns and apply them to new inputs, effectively making use of its pre-trained knowledge for task-specific reasoning. ICL offers several advantages. First, since the examples are written in natural language, it provides an intuitive and interpretable way to interact with large language models, enhancing the flexibility of task definition. Second, ICL does not rely on traditional supervised fine-tuning, meaning that the model's parameters do not need to be modified when adapting to new tasks, significantly reducing computational costs. Furthermore, ICL is applicable to a wide range of tasks, including text generation, logical reasoning, code generation, and optimization problem solving, thereby enhancing the generalization ability of large language models across diverse application scenarios.
\subsection{Prompt Engineering}
\label{2.3}
Prompt engineering is the process of designing and refining the input queries, or "prompts," given to LLMs to elicit specific, desired responses. By carefully crafting prompts, developers and information seekers can guide LLMs to generate more accurate, relevant, and useful outputs. This approach enhances the performance of LLMs, enabling them to solve a wide range of problems in various domains more efficiently. Unlike traditional model fine-tuning, prompt engineering does not require modifying the model’s parameters, making it a more cost-effective solution for adapting models to new tasks. Additionally, prompt engineering supports the exploration of new applications, such as more effective conversational AI systems, chatbots, and virtual assistants, by offering customizable programming that can shape the model's responses and interactions. Prompt engineering involves several key techniques to optimize the input-output interaction of language models. The few-shot learning technique helps the model understand and adapt to new tasks by providing a small number of examples. The prompt chaining method breaks complex tasks into smaller steps, with each step's output serving as the input for the next, ensuring task coherence. Prompt fine-tuning adjusts the structure and content of the prompt to improve model performance. Lastly, adaptive prompts dynamically adjust instructions based on real-time feedback, enhancing the model's flexibility and adaptability. These techniques significantly improve the effectiveness of prompt engineering, enhancing the interaction and task handling capabilities of large language models. As the field evolves, prompt engineering is poised to be increasingly integral to unlocking the full potential of LLMs, providing valuable tools for developers and creating new opportunities in AI research and application.

\section{Methodology}
This section introduces the LLM-RBDO framework, a generative evolutionary optimization method based on the in-context learning capability of large language models. Subsection \ref{3.1} provides an overview of the problem formulation addressed by LLM-RBDO. Subsection \ref{3.2} introduces the process of reliability analysis. Subsection \ref{3.3} discusses the design of prompts. Subsection \ref{3.4} outlines the complete workflow of the LLM-RBDO method, while Subsection \ref{3.5} describes the generation strategy, parameter settings, and convergence conditions of the algorithm.
\subsection{Problem Description}
\label{3.1}
The goal of LLM-RBDO is to identify the optimal system design that minimizes design costs while satisfying reliability requirements under system uncertainties. This problem can be formulated as:
\begin{equation}
\begin{array}{ll}
\text { Minimize: } & \operatorname{Cost}(\mathbf{d}) \\
\text { subject to: } & \operatorname{Pr}\left(G_i(\mathbf{x}, \mathbf{d})<0\right) \leq 1-\Phi\left(\beta_{t_i}\right), \quad i=1, \ldots, n c \\
& \mathbf{d}^{\mathbf{L}} \leq \mathbf{d} \leq \mathbf{d}^{\mathbf{U}}, \quad \mathbf{d} \in R^{\mathrm{nd}} \quad \text { and } \quad \mathbf{x} \in R^{\mathrm{nr}}
\end{array}
\end{equation}
where $\operatorname{Pr}\left(G_i(\mathbf{x}, \mathbf{d})<0\right)$ represents the failure probability, $\beta_{t}$ is the target reliability index, $\operatorname{Cost}(\mathbf{d})$ is the objective function, and $\mathbf{d}$ is the design vector, which includes the mean values of the random design variables x and the deterministic design variables. Additionally, $nc$ , $nd$, and $nr$ represent the number of probability constraints, the number of design variables, and the number of random variables, respectively.
\subsection{Reliability Analysis}
\label{3.2}
This paper constructs a Kriging surrogate model for reliability analysis. The Kriging surrogate model is a non-parametric interpolation model based on statistical principles. To predict the value at a certain point, the model estimates the unknown information at that point by performing a linear combination of the known information within a specified range around it. This estimation process is not influenced by random errors. The Kriging surrogate model exhibits excellent global convergence and high fitting accuracy, demonstrating strong nonlinear approximation capabilities. This means the model can provide accurate predictions when handling complex relationships and nonlinear mappings, and it performs excellently across the entire domain. Its performance function is typically given by the following formula:
\begin{equation}
G_K(\mathbf{x})=\mu+S(\mathbf{x})
\end{equation}
where $\mathbf{x}$ represents the n-dimensional design variables, $G_k(\mathbf{x})$ is the Kriging model's estimation of the performance function at the sample point $\mathbf{x}$, $\mu$ is the global simulation approximation of the sample space, and $S(\mathbf{x})$ is the Gaussian stochastic process with a mean of 0 and a variance of $\sigma^2$. The covariance function of $S(\mathbf{x})$ is given by:
\begin{equation}
\operatorname{Cov}\left[S\left(\mathbf{x}_i\right), S\left(\mathbf{x}_j\right)\right]=\sigma^2 \mathbf{R}\left(\mathbf{x}_i, \mathbf{x}_j\right)
\end{equation}
where $\mathbf{R}$ denotes the correlation matrix; $\mathbf{R}\left(\mathbf{x}_i, \mathbf{x}_j\right)$ can be defined based on the distance between two sample points $\mathbf{x}_i$ and $\mathbf{x}_j$ expressed as:
\begin{equation}
\mathbf{R}\left(\mathbf{x}_i, \mathbf{x}_j\right)=\exp \left[-d\left(\mathbf{x}_i, \mathbf{x}_j\right)\right]
\end{equation}
\begin{equation}
d\left(\mathbf{x}_i, \mathbf{x}_j\right)=\sum_{p=1}^k a_p\left|x_i^p-x_j^p\right|^{b_p}
\end{equation}
where $\mathbf{x}_i$ and $\mathbf{x}_j$ represent two sample points, and $a_p$ and $b_p$ are the parameters of the Kriging model.
For a given point $\mathbf{x}^\prime$, the response value and the mean squared error of the Kriging model are given by:
\begin{equation}
G_K\left(\mathbf{x}^{\prime}\right)=\mu+\mathbf{r}^T \mathbf{R}^{-1}(\mathbf{G}-\mathbf{A} \mu)
\end{equation}
\begin{equation}
e\left(\mathbf{x}^{\prime}\right)=\sigma^2\left[1-\mathbf{r}^T \mathbf{R}^{-1} \mathbf{r}+\frac{\left(1-\mathbf{A}^T \mathbf{R}^{-1} \mathbf{r}\right)^2}{\mathbf{A}^T \mathbf{R}^{-1} \mathbf{A}}\right]
\end{equation}
where $\mathbf{r}$ is the correlation vector between $\mathbf{x}^\prime$ and the sample points $\mathbf{x}_1\sim\mathbf{x}_n$, and $\mathbf{A}$ is an $n\times1$ column vector. The parameters $\mu$ and $\sigma^2$ are given by:
\begin{equation}
\mu=\left[\mathbf{A}^T \mathbf{R}^{-1} \mathbf{A}\right]^{-1} \mathbf{A}^T \mathbf{R}^{-1} \mathbf{G}
\end{equation}
\begin{equation}
\sigma^2=\frac{(\mathbf{G}-\mathbf{A} \mu)^T \mathbf{R}^{-1}(\mathbf{G}-\mathbf{A} \mu)}{n}
\end{equation}
After constructing the Kriging model, this study employs the Monte Carlo simulation for reliability analysis. First, $N$ Monte Carlo samples are generated based on the randomness of the input variables, denoted as \(\mathbf{X}_m \left(\mathbf{x}_{m,1}, \mathbf{x}_{m,2}, \ldots, \mathbf{x}_{m,i}, \ldots, \mathbf{x}_{m,N}\right)\), where \(\mathbf{x}_{m,i}\) represents the $ith$ sample point. Consequently, the response value at this point is estimated by the Kriging model as a normally distributed random variable following \(G\left(\mathbf{x}_{m,i}\right) \sim N\left(G_k(\mathbf{x}_{m,i}), e(\mathbf{x}_{m,i})\right)\). Therefore, \(\mathbf{x}_{m,i}\) can be classified as either failure ("1") or safe ("0"), defined as:
\begin{equation}
I_f\left(\mathbf{x}_{m,i}\right) =
\begin{cases}\label{eq11} 
1, & \text{if } G_k\left(\mathbf{x}_{m,i}\right) < 0 \\
0, & \text{otherwise}
\end{cases}
\end{equation}
The reliability of the point can then be represented as:
\begin{equation}\label{eq12}
R=1-\frac{\sum_{i=1}^N I_f\left(\mathbf{x}_{m, i}\right)}{N}
\end{equation}
\subsection{Prompt Design}
\label{3.3}
\begin{figure}
    \centering
    \includegraphics[width=0.75\linewidth]{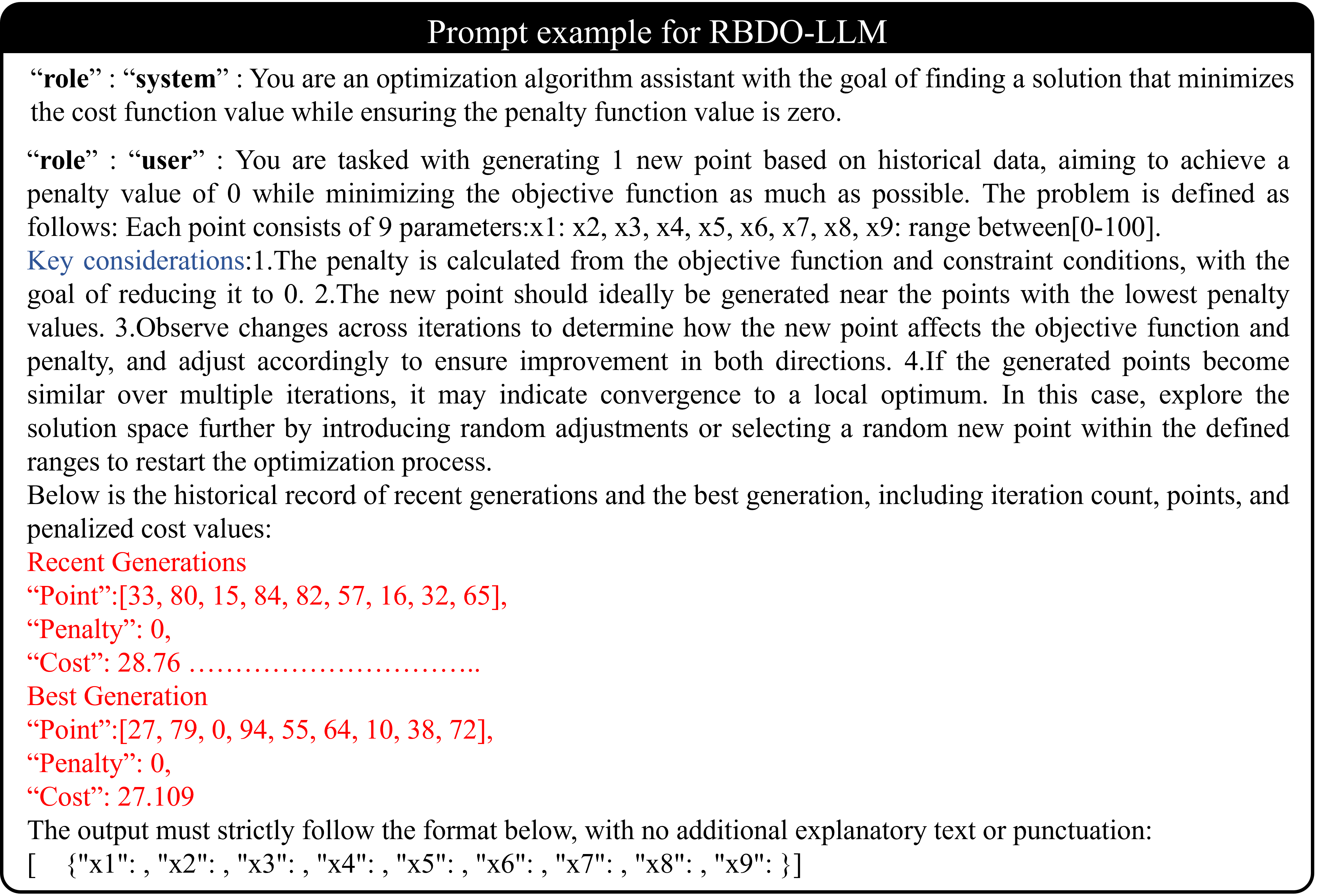}
    \caption{Example of LLM-RBDO Prompt}
    \label{fig:2}
\end{figure}

Prompt design is a crucial element in interacting with large language models, guiding the model to generate accurate and relevant outputs through simple textual input. Well-crafted prompts enable the model to quickly understand the task's objectives, reduce trial and error, and enhance efficiency. In particular, for complex tasks, a well-designed prompt not only assists the model in performing step-by-step reasoning but also allows for better control over the direction and style of the output. Furthermore, prompts enable the model to adapt to the specific requirements of various scenarios and fields, facilitating the rapid generation of results that meet professional standards. This paper designs a few-shot learning prompt architecture for the LLM-RBDO framework, with a specific example shown in Figure \ref{fig:2}. First, the LLM is assigned the role of an optimization algorithm assistant, generating new design points $\mathbf{x}^{(t+1)}=\left[x_1, x_2, \cdots, x_n\right]$ that minimize the cost function while maintaining a penalty function value of zero (see Subsection \ref{3.4} for the penalty function definition). The generated points must remain within the feasible design space: 
\begin{equation}
\mathbf{x}^{(t+1)} \in \mathbb{X}, \quad \mathbb{X}=\left[x_1^{\min }, x_1^{\max }\right] \times \cdots \times\left[x_n^{\min }, x_n^{\max }\right]
\end{equation}
The next step involves providing the model with the design variable ranges and some guiding information to steer the optimization direction. The specific guidance includes explaining the role of the penalty function and directing the model to optimize for solutions where the penalty function value approaches zero and the cost function value decreases. To guide the search, LLM is provided with key historical optimization data, including the most recent $K$ iterations of design points, their corresponding cost function values, and penalty function values:
\begin{equation}
\mathbf{H}^{(t)}=\left\{\left(\mathbf{x}^{(t-k)}, \operatorname{Cost}\left(\mathbf{x}^{(t-k)}\right), \text { Penalty }\left(\mathbf{x}^{(t-k)}\right)\right)\right\}_{k=1}^K
\end{equation}
where $\mathbf{x}^{(t-k)}$ represents the design point at generation $(t-k)$, $\operatorname{Cost}\left(\mathbf{x}^{(t-k)}\right)$ denotes the cost function value of the design point at generation $(t-k)$, and $\text { Penalty }\left(\mathbf{x}^{(t-k)}\right)$ represents the penalty function value of the design point at generation $(t-k)$. This historical information enables the model to infer optimization trends and refine its point-generation strategy accordingly.

A structured output format is enforced to ensure consistency and usability in subsequent optimization steps. The model is required to output the next design point in a strict JSON format:$\text { [ \{"x1": value, "x2": value, ..., "xn": value \} ] }$. This constraint guarantees machine-readability and facilitates seamless integration with the iterative optimization process. To avoid premature convergence to local optima, the prompt also incorporates an adaptive generation strategy. The model is instructed to prioritize generating new points near those with the lowest penalty values while continuously evaluating how changes affect both cost function and penalty function. If consecutive iterations produce highly similar solutions, indicating possible stagnation, the model is guided to introduce greater diversity in its generated points. This can involve adjusting the search region or incorporating additional randomness to explore new areas of the design space, thereby enhancing the robustness of the optimization process. By integrating structured input constraints, historical iteration data, and an adaptive point-generation mechanism, the LLM effectively participates in the optimization process, leveraging its in-context learning capability to iteratively refine design points and improve optimization efficiency.
\subsection{LLM-Based Generative Design}
\label{3.4}
\begin{figure}
    \centering
    \includegraphics[width=0.75\linewidth]{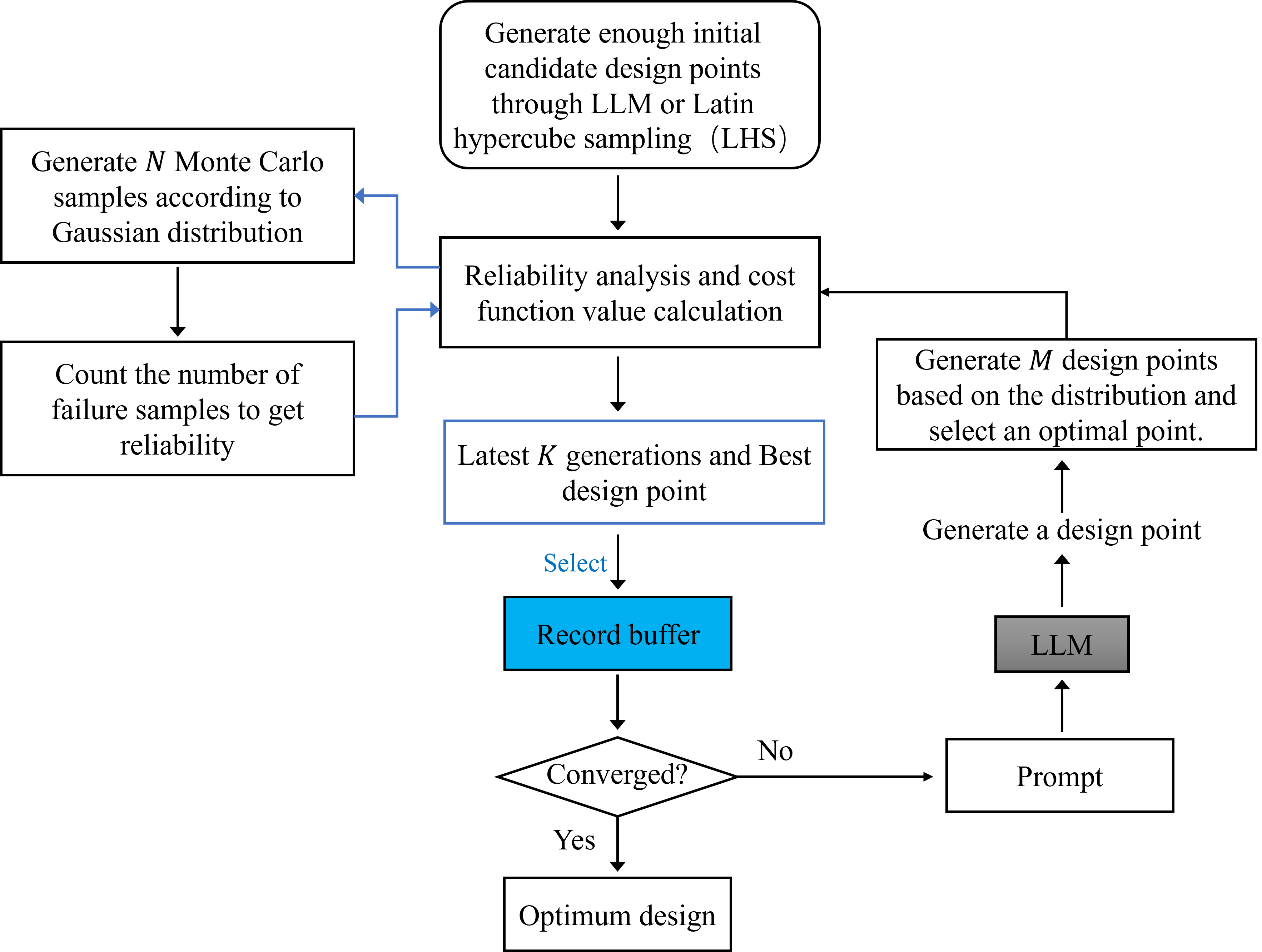}
    \caption{Overall Flowchart of LLM-RBDO}
    \label{fig:3}
\end{figure}

Similar to evolutionary strategies, LLM-RBDO generates new points $x_t$ iteratively to explore the optimization direction and identify new design points. The detailed workflow is illustrated in Figure \ref{fig:3}. First, $N$ candidate initial design points are initialized, denoted as \(\mathbf{X}_d \left(\mathbf{x}_{d,1}, \mathbf{x}_{d,2}, \ldots, \mathbf{x}_{d,i}, \ldots, \mathbf{x}_{d,N}\right)\), where  \(\mathbf{x}_{d,i}\) represents the $i$-th initial sample point. These $N$ initial design points are randomly and uniformly sampled within the range of design variables. Next, the cost function value $\operatorname{Cost}(\mathbf{X_d})$ is evaluated, and reliability analysis is conducted using Monte Carlo simulation. This study assumes that all random variables follow an n-dimensional Gaussian distribution with a mean of $\overline{\mathbf{X}}$ and a fixed variance matrix  $\sigma^2 \mathbf{I}$, expressed as:
\begin{equation}
\mathbf{x}_{d, i} \sim N\left(\overline{\mathbf{X}}, \sigma^2 \mathbf{I}\right)
\end{equation}
Based on the randomness of the input variables, $N$ Monte Carlo samples can be generated. The reliability corresponding to each performance function for each design point can then be calculated using equations (\ref{eq11}) and (\ref{eq12}).

Since RBDO problems typically involve multiple constraint functions, when the LLM receives too much information, it may lead to a decrease in its reasoning ability. Additionally, all LLMs have an input length limitation. To address this, we use a penalty function to represent whether the constraints meet the target reliability requirements. The penalty function is defined as follows:
\begin{equation}
\text { Penalty }=\sum_{i=1}^n \begin{cases}w_p \cdot\left(R_t-R_i\right)^2, & \text { if } R_i<R_t \\ 0, & \text { if } R_i \geq R_t\end{cases}
\end{equation}
where $\text { Penalty }$  represents the penalty function value, $R_t$ is the target reliability value, $R_i$ is the reliability value corresponding to the $i$-th performance function, $w_p$ is the penalty coefficient, and n is the total number of reliability constraints.

By evaluating the cost function and penalty values for each candidate initial design point, the points are filtered to obtain the first generation of design points, which are then stored in the buffer. The filtering criteria are as follows: if the penalty function values for all candidate initial design points are greater than or equal to 0, it indicates that none of the points meet the reliability design requirements. In this case, the design point with the smallest penalty function value is selected as the first generation design point. If there are design points with penalty function values equal to 0, these points satisfy the reliability design requirements. Among these points, the one with the smallest cost function value is selected as the first generation design point.

Subsequently, the LLM guides the search for the optimal design point. First, a prompt containing information about the initial design points is prepared. Based on this prompt, the LLM generates the next-generation design point. Then, $M$ additional design points are generated around the newly obtained design point according to a given Gaussian distribution. Since some of these $M$ design points may fall outside the allowable range of design variables, a filtering process is applied to remove infeasible points. The cost function and penalty function values of the filtered design points are then evaluated. Following the same selection criteria as in the initial stage, the best design point is chosen from the filtered candidates. The newly selected design point is stored in the buffer along with its corresponding penalty function value and cost function value. This updated buffer information is used to refine the prompt, guiding the LLM in generating the next set of design points. This iterative process continues until the convergence criteria are met.
\subsection{Generation Strategy and Parameter Settings}
\label{3.5}
In this paper, Deepseek-V3 \cite{liu2024deepseek} is selected to implement the LLM-RBDO framework. During the design optimization process, when the design variables have high dimensionality or when multiple constraints are present, the LLM may not find a solution that satisfies all constraints during multiple iterations searching for the optimal solution. However, it might encounter a design point with the lowest cost function value in history, but with a small penalty function value. To allow the large model to continue iterating and searching for the next optimal solution, this point can be considered as the historical optimal design point and stored in the Record buffer. Therefore, the criteria for determining the optimal design point are set as follows: the cost function value is lower than the historical optimal cost function value, and the penalty function value is below a user-defined threshold $\delta$. The convergence condition for the entire LLM-RBDO framework is either when the optimal design point does not update for $T$ consecutive iterations, or when the maximum number of iterations $S$ is reached. Since LLMs exhibit certain limitations when handling floating-point numbers, this paper maps the coordinates of design points in the input prompts to the range [0, 100]. The randomness and diversity of LLM-generated content can generally be controlled by adjusting the parameters $\text{top\_p}$ and temperature. The temperature parameter primarily regulates the randomness of the model’s predictions. A higher temperature increases output diversity and creativity by raising the likelihood of selecting lower-probability words, making it suitable for open-ended or creative tasks. Conversely, a lower temperature makes the model more deterministic, favoring high-probability words and producing more focused and predictable outputs. The $\text{top\_p}$ parameter dynamically adjusts the size of the candidate word set based on cumulative probability distribution. A higher $\text{top\_p}$ allows the model to sample from a broader range, enhancing content diversity, while a lower $\text{top\_p}$ significantly reduces randomness, generating more concentrated outputs. In design optimization tasks, it is crucial to ensure that the LLM explores a sufficiently broad range of optimization directions and selects the most promising ones for further investigation. Therefore, in this study, $\text{top\_p}$ is set to 0.9 to introduce sufficient randomness in generating new design points, while temperature is set to 0.2 to ensure the selection of the most probable design points for the next-generation optimization process.

\section{Case studies}
In this section, three case studies are presented to demonstrate the application of the developed generative reliability-based design optimization method (LLM-RBDO) based on the contextual learning capabilities of large language models. These case studies include a two-dimensional mathematical design problem and a high-dimensional vehicle side crash design problem.
\subsection{Two-Dimensional Mathematical Design Problem}
\begin{figure}
    \centering
    \includegraphics[width=0.5\linewidth]{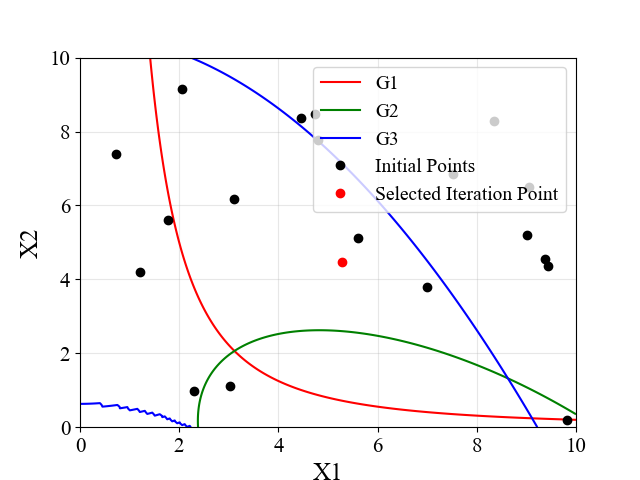}
    \caption{Contour of the Limit State Function and Candidate Initial Design Points}
    \label{fig:4}
\end{figure}

\begin{figure}
    \centering
    \includegraphics[width=0.5\linewidth]{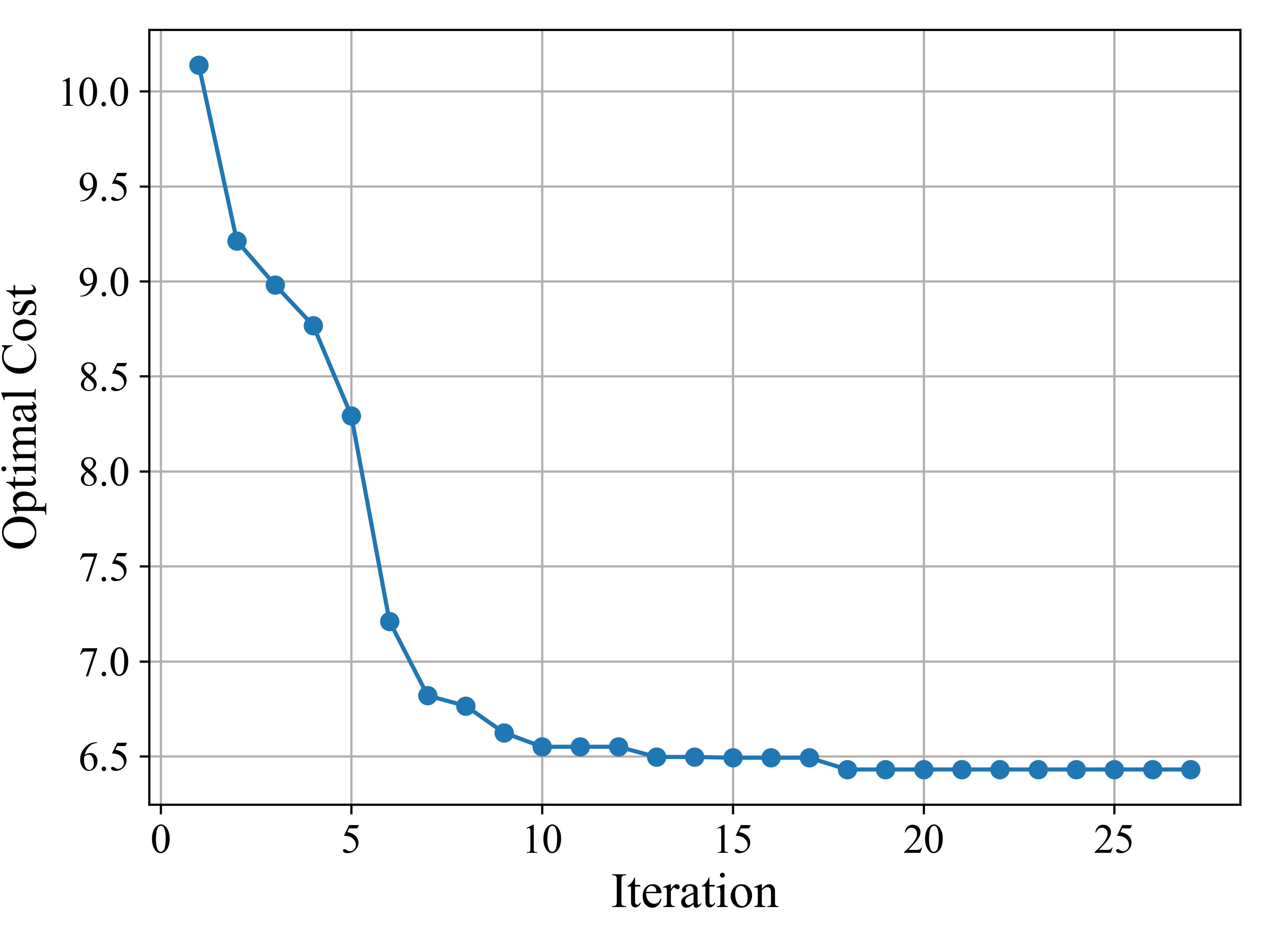}
    \caption{Iterative Process of Optimal Cost Function Value in LLM-RBDO (Case Study 1)}
    \label{fig:5}
\end{figure}

\begin{figure}
    \centering
    \includegraphics[width=0.5\linewidth]{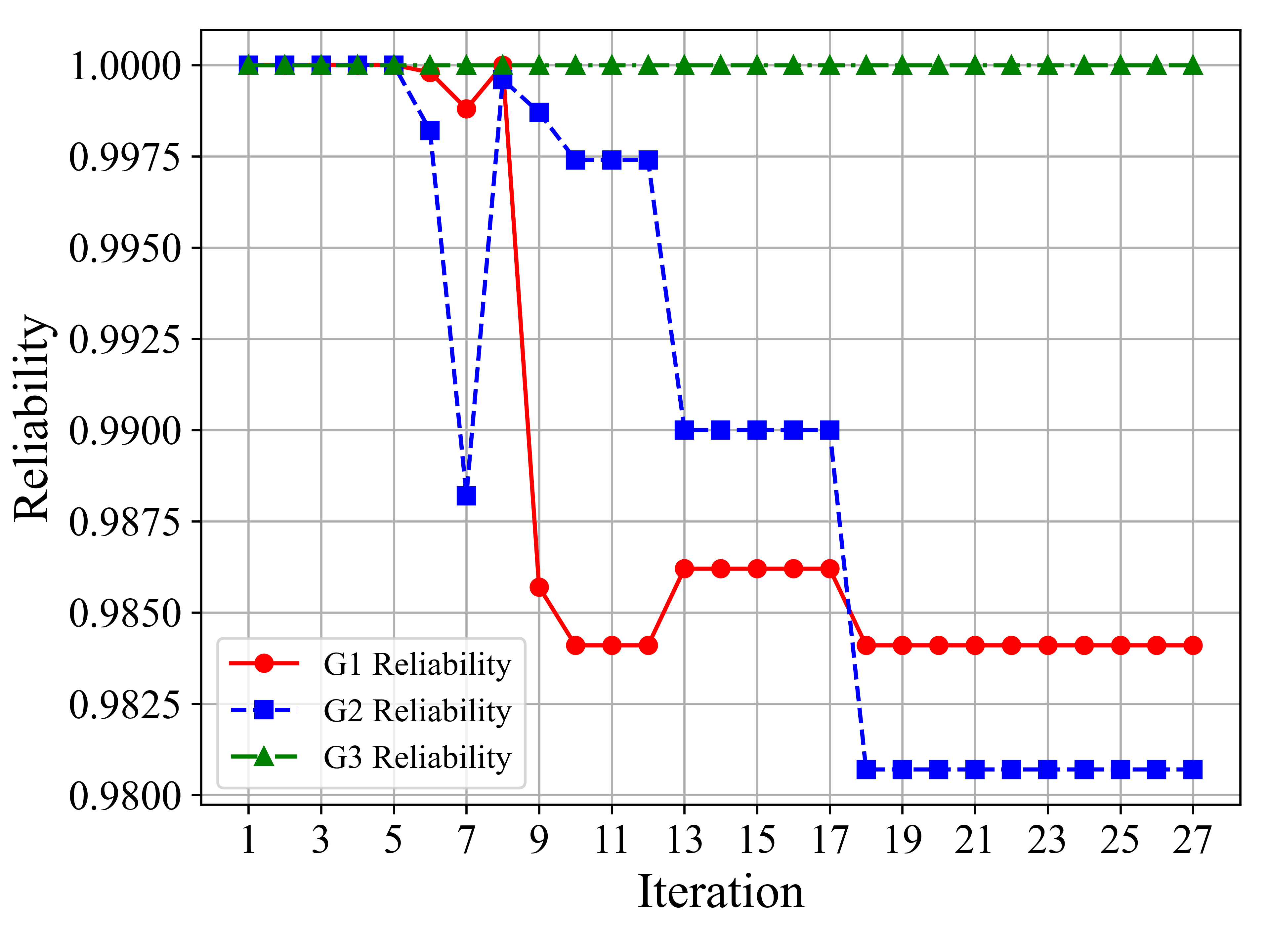}
    \caption{Iteration Process of Design Point Reliability (Case Study 1)}
    \label{fig:6}
\end{figure}

\begin{figure}
    \centering
    \includegraphics[width=0.5\linewidth]{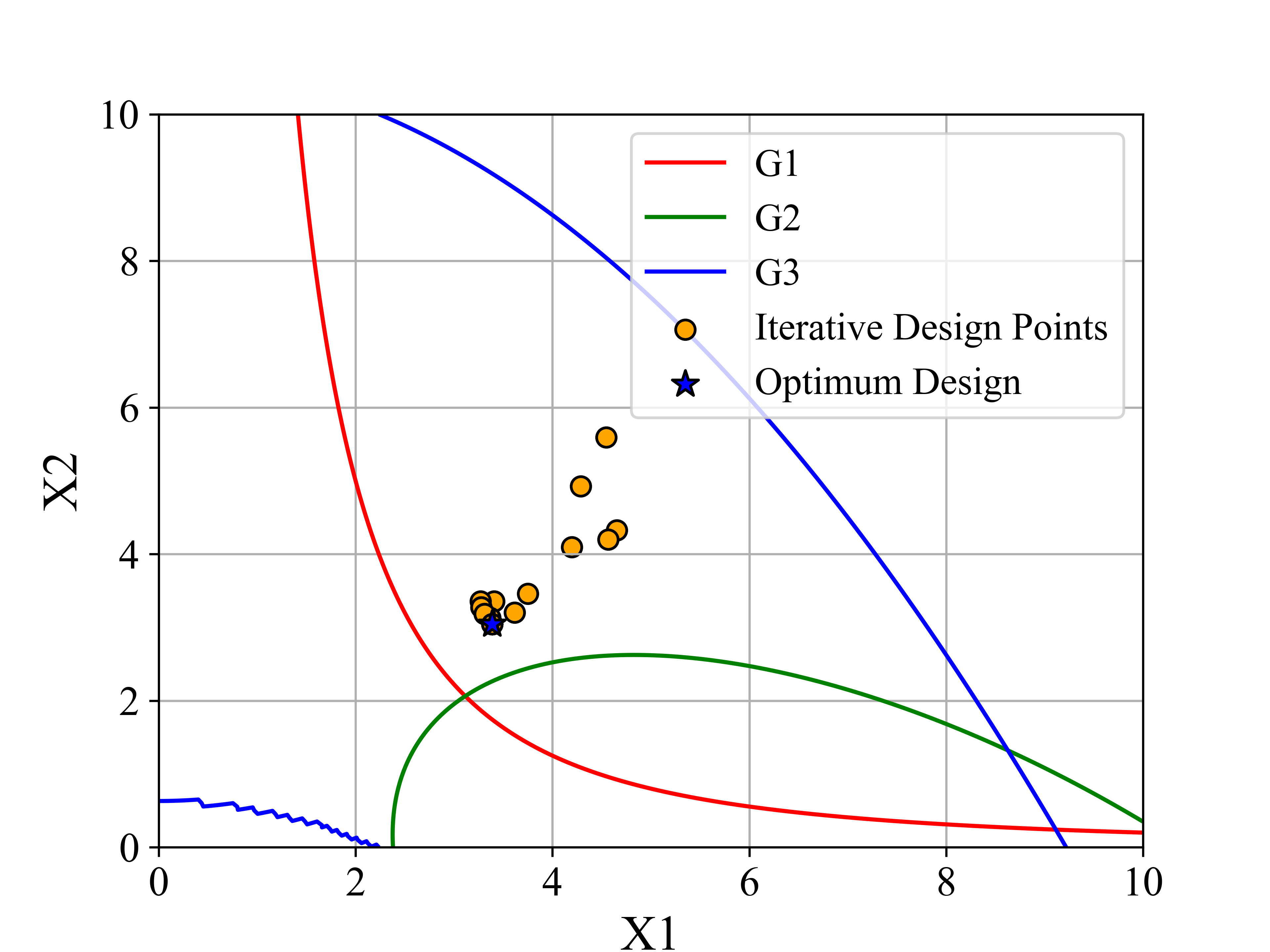}
    \caption{Iteration Process of Design Points}
    \label{fig:7}
\end{figure}

\begin{figure}
    \centering
    \includegraphics[width=0.5\linewidth]{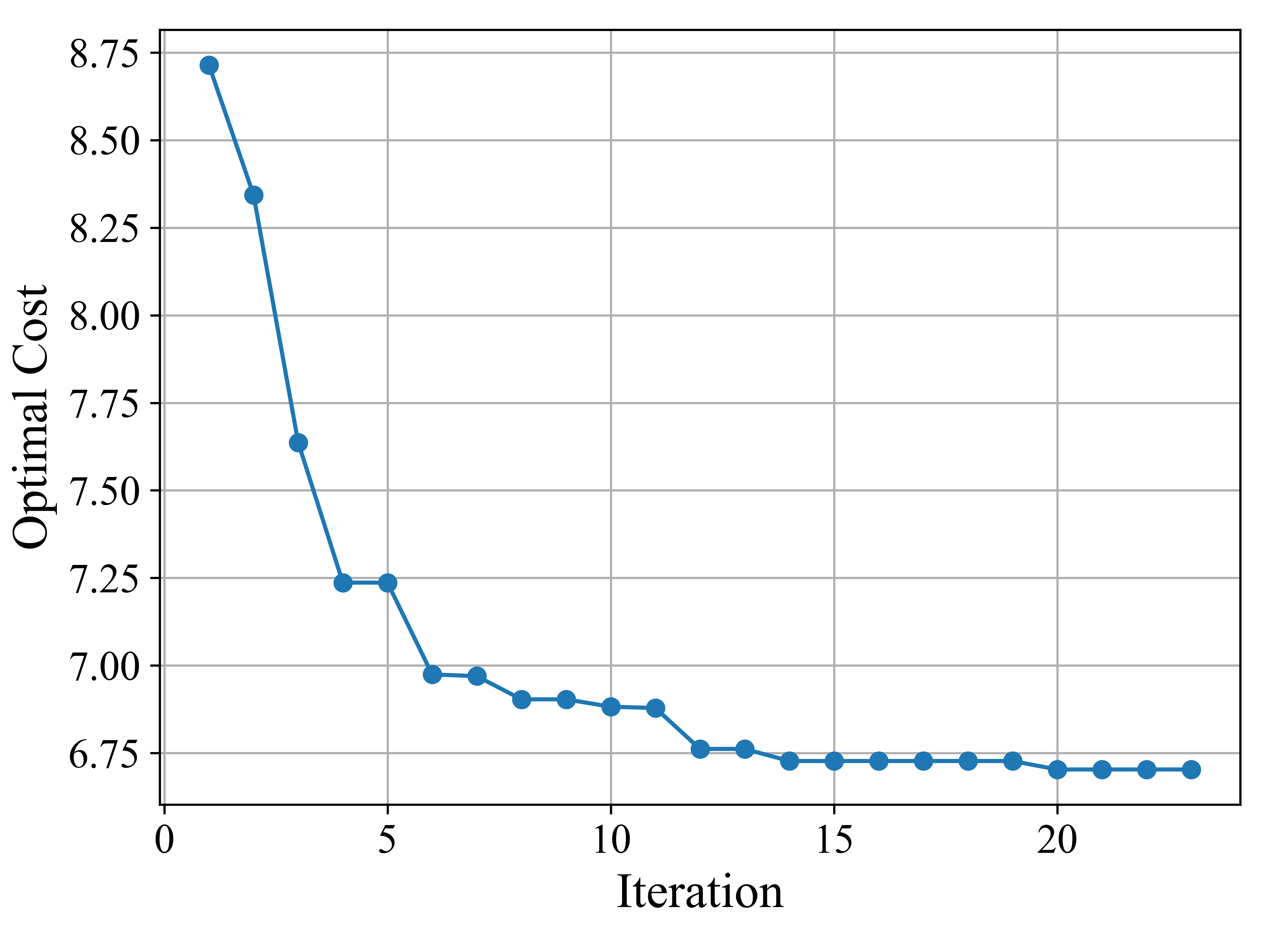}
    \caption{Iterative Process of Optimal Cost Function Value in SEGA (Case Study 1)}
    \label{fig:8}
\end{figure}

The mathematical case involves an optimization problem with two random design variables, $X_1$ and $X_2$, both of which follow a normal distribution. Specifically, $X_1 \sim N\left(\mu_1, 0.3464^2\right)$ and $X_2 \sim N\left(\mu_2, 0.3464^2\right)$, where the design vector is given by $\mathbf{d}=\left[d_1, d_2\right]^T=\left[\mu\left(X_1\right), \mu\left(X_2\right)\right]^T$.The target reliability value $R_t$ is set to 0.98. Therefore, the RBDO problem is formulated as follows:
\begin{equation}
\begin{array}{ll}
\text { Minimize: } & \text { Cost }=d_1+d_2 \\
\text { subject to: } & \operatorname{Pr}\left[G_i(X)<0\right] \leq 1-R_t, i=1 \sim 3 \\
& 0 \leq d_1 \& d_2 \leq 10 \\
& G_1=X_1^2 X_2 / 20-1 \\
& G_2=\left(X_1+X_2-5\right)^2 / 30 \\
& \quad+\left(X_1-X_2-12\right)^2 / 120-1 \\
& G_3=80 /\left(X_1^2+8 X_2-5\right)-1
\end{array}
\end{equation}
In this case study, the maximum number of iterations is set to 50, with the convergence criterion defined as no updates to the optimal design point for 10 consecutive iterations. Each time the LLM generates a recommended design point based on a given prompt, 10 additional design points are sampled around it following a Gaussian distribution. The optimal design point is determined using a penalty function, which must be below a predefined threshold of $\delta=0.01$. Additionally, information from the five most recent generations is retained and provided to the LLM for reference. According to the workflow of LLM-RBDO, 20 sample points are first selected using Latin Hypercube Sampling, and 3 Kriging models ($\mathbf{M_1}$, $\mathbf{M_2}$, $\mathbf{M_3}$) are constructed. The number of candidate initial design points to be generated is set to 20. Due to the randomness of the design points generated in each experiment, Figure \ref{fig:4} illustrates the distribution of the generated points along with the contour of the limit state function for a particular trial. The black points represent the generated candidate initial design points, while the red point indicates the selected first-generation iteration point $\mathbf{d_0}=[5.28,4.47]$. Subsequently, the LLM generates the next-generation iteration points based on the provided first-generation iteration point. 
The iterative process of the optimal cost function is shown in Figure \ref{fig:5}. It can be observed that LLM-RBDO achieves the optimal solution at the 18th generation and terminates at the 27th generation. The obtained optimal solution is $\mathbf{d_llmopt}=[3.38,3.04]$, with a corresponding cost function value of 6.431. At this point, the reliability values for $G_1$, $G_2$, and $G_3$ are 0.9841, 0.9807, and 1, respectively. Based on the specified distribution, 10,000 real sample points are generated using Monte Carlo simulation for reliability analysis. The results shows that the reliability values obtained are: $G_1 = 0.9834$, $G_2 = 0.9805$, and $G_3 = 1$. These values closely align with those predicted by the Kriging model, confirming the accuracy of the results obtained through the LLM-RBDO method. The iteration of the reliability of the optimal design point in each generation is shown in Figure \ref{fig:6}. It can be observed that the reliability of $G_3$ remains constant at 1, while the reliability of $G_1$ and $G_2$ fluctuates slightly in the middle, but overall shows a downward trend until convergence. The iteration process of the design points is shown in Figure \ref{fig:7}. The yellow points represent the design iteration points, and the blue pentagrams represent the design points in the final generation. It can be seen that these iteration points start far from the optimal design point and gradually approach it as the iterations proceed. 
For comparison, a single-objective strengthen elitist GA algorithm (SEGA) is also applied to this case study. The initial population size is set to 50, with a maximum iteration limit of 50. The results are illustrated in Figure \ref{fig:8}. The SEGA converges at the 23th generation, yielding an optimal solution of $\mathbf{d_gaopt}=[3.12,3.58]$, with a cost function value of 6.703. The corresponding reliability values for $G_1$, $G_2$, and $G_3$ at this solution are 0.9787, 1, and 1, respectively. Through comparison, it is found that LLM-RBDO has a convergence speed similar to SEGA. Moreover, by sacrificing the reliability of $G_2$, LLM-RBDO achieves a solution with a lower cost function value than the optimal solution obtained by SEGA, demonstrating better performance in this case. Furthermore, this study evaluates the performance of LLM-RBDO by comparing its results with those obtained using SEGA when the real function is directly used as a constraint, without employing the Kriging model. Additionally, a comparison with the traditional FORM method is conducted to further validate the effectiveness of LLM-RBDO. As shown in Table \ref{table:1}, the results indicate that the solution obtained by LLM-RBDO achieves the lowest cost function value while satisfying the required reliability constraints among all the compared methods.
\begin{table}[htbp]
    \centering
    \caption{Comparison of Results for Case Study 1}
    \label{table:1}
    \begin{tabular}{l c c c c c c c}
        \toprule
        \textbf{RBDO} & \textbf{Reliability} & \textbf{Optimum} & \textbf{Cost} & \textbf{G$_1$} & \textbf{G$_2$} & \textbf{G$_3$} \\
        \textbf{method} & \textbf{estimation} & \textbf{solution} &  &  &  &  \\
        \midrule
        LLM-RBDO & Kriging & [3.38, 3.04] & 6.43 & 0.9834 & 0.9805 & 1 \\
        LLM-RBDO & True constraint & [3.41, 3.05] & 6.46 & 0.9886 & 0.9813 & 1 \\
        SEGA & Kriging & [3.12, 3.58] & 6.703 & 0.9787 & 1 & 1 \\
        SEGA & True constraint & [3.05, 3.78] & 6.82 & 0.9765 & 1 & 1 \\
        FORM & -- & [3.46, 3.14] & 6.60 & 0.98 & 0.98 & -- \\
        \bottomrule
    \end{tabular}
\end{table}
\subsection{High-Dimensional Vehicle Side Crash Design Problem}
According to vehicle side-impact safety regulations, vehicle designs must meet internal requirements and comply with relevant legal standards. This study adopts the side-impact protection guidelines established by the European Enhanced Vehicle Safety Committee (EEVC), a widely used analysis method. In the EEVC procedure, human dummy responses serve as the primary indicators for evaluating side-impact performance. These include head injury criteria (HIC), abdominal load, pubic symphysis force, viscous criterion (VC), and the deformation of the ribs at the upper, middle, and lower regions. Additionally, the analysis of side impacts also focuses on the speed of the B-pillar midpoint and the velocity of the front door at the B-pillar. The objective of this study is to minimize the overall vehicle weight while ensuring or improving the vehicle's side-impact performance.

The vehicle side-impact design case study involves 11 random variables, including 9 design variables and 2 random parameters. As shown in Table \ref{table:2}, the design variables encompass the thickness of key components ($\mathrm{d_1}$ to $\mathrm{d_7}$) as well as material performance parameters ($\mathrm{d_8}$ and $\mathrm{d_9}$). In addition, the two non-design random parameters are the obstacle height and collision position, with a variation range of ±30 mm (based on actual physical testing data). All random variables are assumed to follow a normal distribution.
\begin{table}[htbp]
    \centering
    \caption{Properties of design and random parameters of vehicle side impact}
    \label{table:2}
    \begin{tabular}{l c c c c c c}
        \toprule
        \textbf{Random variables} & \textbf{Distribution type} & \textbf{Standard deviation} & $\mathrm{d}_{\mathrm{L}}$ & $\mathrm{d}$ & $\mathrm{d}_{\mathrm{U}}$ \\
        \midrule
        $\mathrm{X}_1$ \text{(B-pillar inner)} & Normal & 0.030 & 0.500 & $\mathrm{d_1}$ & 1.500 \\
        $\mathrm{X}_2$ \text{(B-pillar reinforce)} & Normal & 0.030 & 0.500 & $\mathrm{d_2}$ & 1.500 \\
        $\mathrm{X}_3$ \text{(Floor side inner)} & Normal & 0.030 & 0.500 & $\mathrm{d_3}$ & 1.500 \\
        $\mathrm{X}_4$ \text{(Cross member)} & Normal & 0.030 & 0.500 & $\mathrm{d_4}$ & 1.500 \\
        $\mathrm{X}_5$ \text{(Door beam)} & Normal & 0.030 & 0.500 & $\mathrm{d_5}$ & 1.500 \\
        $\mathrm{X}_6$ \text{(Door belt line)} & Normal & 0.030 & 0.500 & $\mathrm{d_6}$ & 1.500 \\
        $\mathrm{X}_7$ \text{(Roof rail)} & Normal & 0.030 & 0.500 & $\mathrm{d_7}$ & 1.500 \\
        $\mathrm{X}_8$ \text{(Mat. B-pillar inner)} & Normal & 0.006 & 0.192 & $\mathrm{d_8}$ & 0.345 \\
        $\mathrm{X}_9$ \text{(Mat. floor side inner)} & Normal & 0.006 & 0.192 & $\mathrm{d_9}$ & 0.345 \\
        $\mathrm{X}_{10}$ \text{(Barrier height)} & Normal & 10 & - & 0 & - \\
        $\mathrm{X}_{11}$ \text{(Barrier hitting)} & Normal & 10 & - & 0 & - \\
        \bottomrule
    \end{tabular}
\end{table}
The RBDO (Reliability-Based Design Optimization) problem for the vehicle side-impact design case can be expressed as:
\begin{equation}
\begin{array}{ll}
\text { Minimize: } & \operatorname{Cost}(\mathbf{d}) \\
\text { subject to: } & P_{v r}\left[G_i(\mathbf{x})<0\right] \leq 1-R_t, i=1 \sim 10 \\
& \mathbf{d}_L \leq \mathbf{d} \leq \mathbf{d}_U, \mathbf{d} \in R^9 \quad \text { and } \quad \mathbf{x} \in R^{11}
\end{array}
\end{equation}
where the cost function $\operatorname{Cost}(\mathbf{d})$ represents the weight and is expressed as:
\begin{equation}
\begin{aligned}
\operatorname{Cost}(\mathbf{d})= & 1.98+4.90 d_1+6.67 d_2+6.98 d_3+4.01 d_4 \\
& +1.78 d_5+2.73 d_7
\end{aligned}
\end{equation}
$R_t$ is set to 0.9, and the constraints $G_1 \sim G_{10}$ correspond to different safety criteria: abdominal load ($G_1$), rib deflections ($G_2 \sim G_4$), viscous criterion (VC) ($G_5 \sim G_7$), pubic symphysis force ($G_8$), B-pillar velocity ($G_9$), and front door velocity ($G_10$). More detailed information on these limit states can be found in the work of Youn et al. \cite{youn2004reliability}.

\begin{figure}
    \centering
    \includegraphics[width=0.5\linewidth]{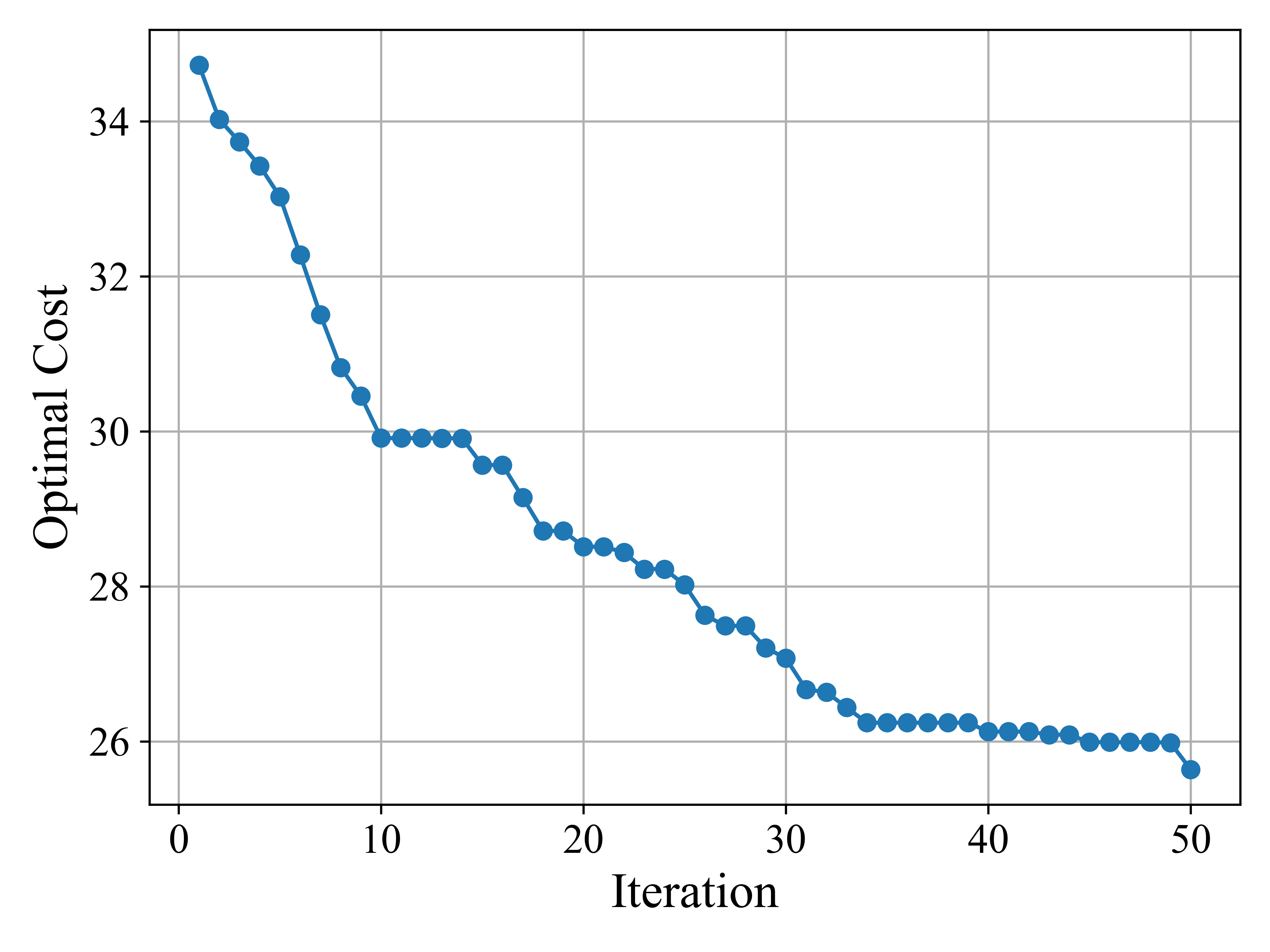}
    \caption{Iterative Process of Optimal Cost Function Value in LLM-RBDO (Case Study 2)}
    \label{fig:9}
\end{figure}

\begin{figure}
    \centering
    \includegraphics[width=0.5\linewidth]{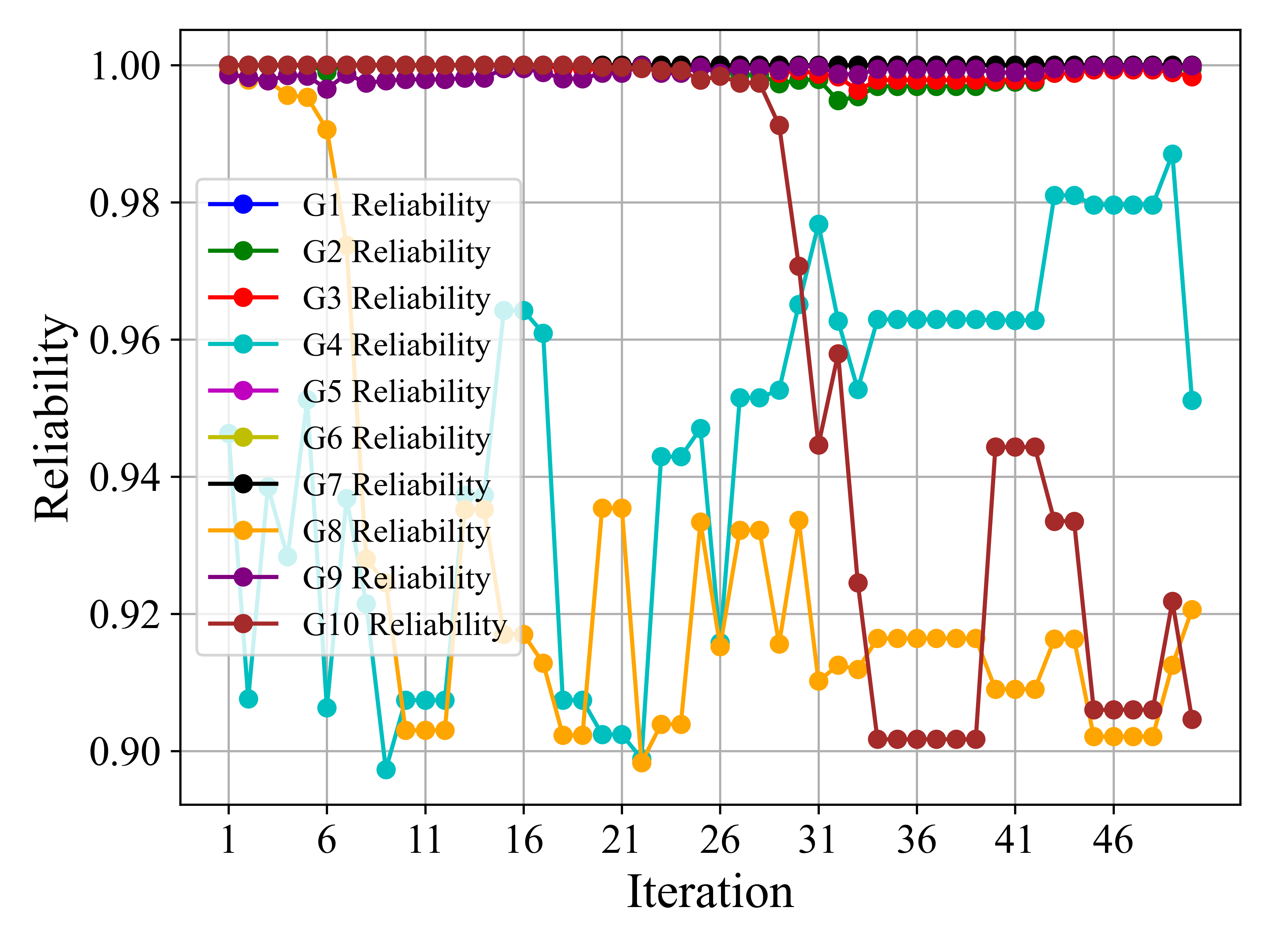}
    \caption{ Iteration Process of Design Point Reliability (Case Study 2)}
    \label{fig:10}
\end{figure}

\begin{figure}
    \centering
    \includegraphics[width=0.5\linewidth]{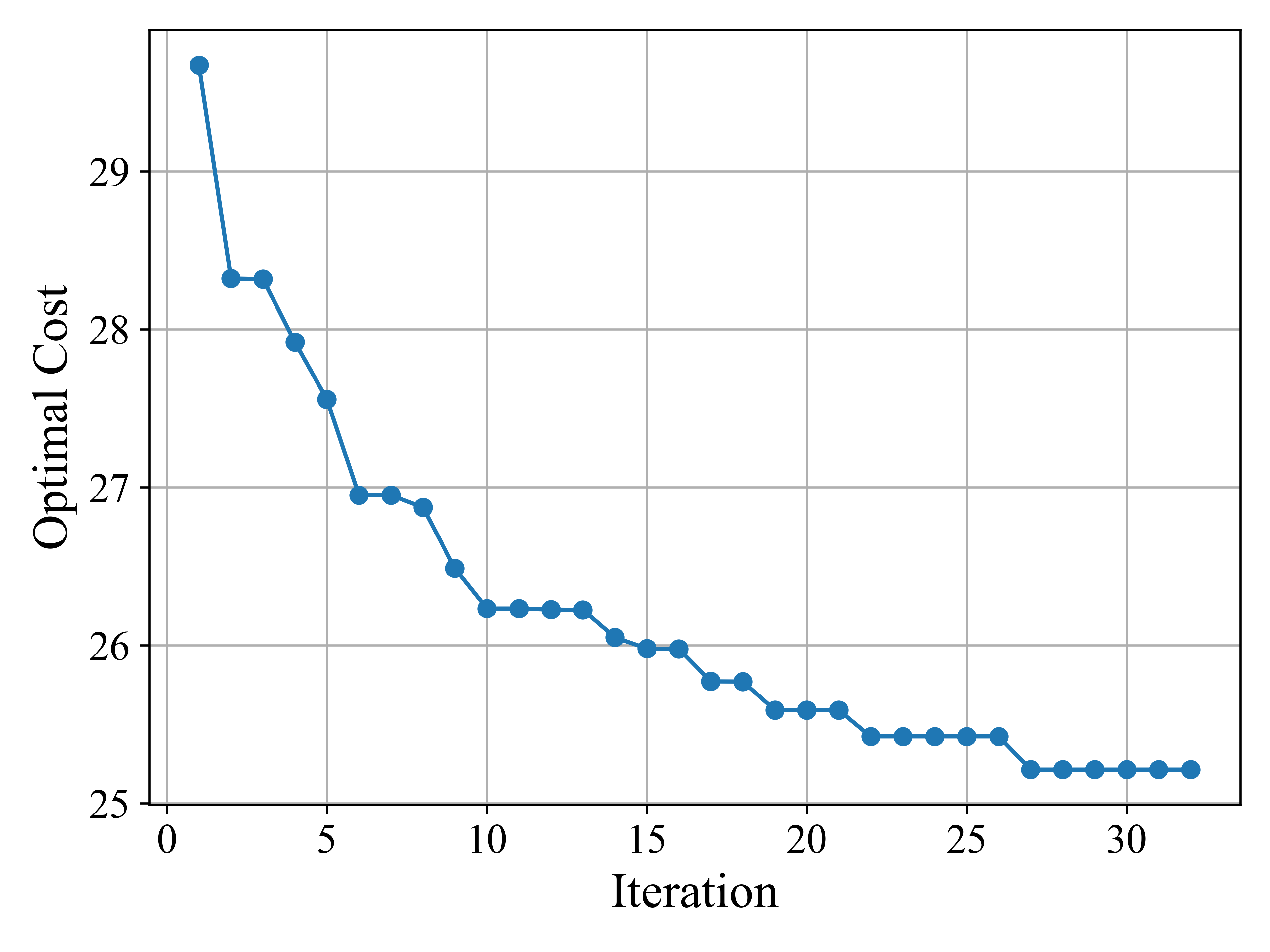}
    \caption{Iterative Process of Optimal Cost Function Value in SEGA (Case Study 2)}
    \label{fig:11}
\end{figure}

To solve the RBDO problem, 100 initial sample points are generated using Latin Hypercube Sampling, and all 10 performance functions are evaluated. Based on the initial dataset D, Kriging models $\mathbf{M_1} \sim \mathbf{M_{10}}$ are constructed for each of the 10 performance functions using the initial samples and their corresponding performance responses. In this case study, the maximum number of iterations is set to 50, with the convergence criterion defined as no updates to the optimal design point for 10 consecutive iterations. The optimal design point is determined based on a penalty function threshold, where a solution is considered optimal if its penalty function value is below a predefined threshold $\delta=0.1$. In each iteration, the LLM generates a recommended design point based on the given prompt and subsequently samples 10 additional design points around it following a Gaussian distribution. The information provided to the LLM includes the coordinates of design points, penalty function values, and cost function values from the most recent five generations. Due to the high randomness inherent in high-dimensional problems, the LLM-RBDO method generates 60 candidate initial design points to ensure the selection of a well-performing first-generation design point. As in case study 1, the application of LLM-RBDO in this study is demonstrated using the results from a specific experiment. The first-generation design point selected by LLM-RBDO is [1.12, 1.29, 1.27, 1.37, 1.16, 1.13, 0.97, 0.29, 0.28], with a corresponding cost function value of 34.72 and a penalty function value of 0. The iterative process of optimizing the cost function is illustrated in Figure \ref{fig:9}. The optimal design point is identified in the 50th generation as [0.52, 1.35, 0.5, 1.38, 0.73, 1.23, 0.58, 0.31, 0.26], achieving a cost function value of 25.59 and a penalty function value of 0. The corresponding reliability values for the constraints are as follows: $G_1=1$, $G_2=0.989$, $G_3=0.9978$, $G_4=0.9545$, $G_5=1$, $G_6=1$, $G_7=1$, $G_8=0.9092$, $G_9=0.9993$, and $G_10=0.9042$. Similarly, by applying the Monte Carlo method to calculate the true reliability of the optimal design point based on the given distribution, the following reliability values are obtained: $G_1=1$, $G_2=0.9995$, $G_3=0.9974$, $G_4=0.9274$, $G_5=1$, $G_6=1$, $G_7=1$, $G_8=0.9047$, $G_9=0.9998$, and $G_10=0.9022$. The reliability results are found to be very close to the optimal solution reliability predicted by the Kriging model during the LLM-RBDO optimization process. The reliability iteration process of the optimal design point is shown in Figure \ref{fig:10}. The reliability of $G_5$, $G_6$, and $G_7$ remains constant at 1, while the reliability of $G_2$ $G_3$, and $G_9$ fluctuates slightly around 1. The reliability of $G_4$, $G_8$, and $G_10$ exhibits larger fluctuations during the iteration process. The iterative process of optimizing the cost function using SEGA is shown in Figure \ref{fig:11}. After 32 iterations, the optimal design point obtained is [0.5, 1.39, 0.5, 1.31, 0.773, 1.44, 0.5, 0.224, 0.34], with a corresponding cost function value of 25.21. The reliability values for the constraints are as follows: $G_1=1$, $G_2=0.8987$, $G_3=0.9697$, $G_4=0.9183$, $G_5=1$, $G_6=1$, $G_7=1.0$, $G_8=0.8937$, $G_9=0.9452$, and $G_10=0.9638$. A comparison shows that while LLM-RBDO does not converge as quickly as in low-dimensional problems, its performance in solving high-dimensional problems with multiple constraints is quite similar to that of the SEGA evolutionary algorithm. This study also evaluates the performance of  LLM-RBDO by computing its results and those of SEGA when the real function is directly used as a constraint without employing the Kriging model. Additionally, a comparison with the traditional FORM method is conducted to further validate its effectiveness. As shown in Table \ref{table:3} and Table \ref{table:4}, the results indicate that for high-dimensional and complex problems, LLM-RBDO fails to achieve an optimal solution comparable to that of the traditional FORM and SEGA methods, instead converging to a near-optimal solution. This issue may be alleviated by increasing the number of iterations, but the fundamental challenge lies in the limitations of large language models when handling highly complex constraints.

\begin{table}[htbp]
    \centering
    \caption{Comparison of results for case study 2(a)}
    \label{table:3}
    \begin{tabular}{l c c c c c c}
        \toprule
        \textbf{RBDO method} & \textbf{Reliability estimation} & \textbf{Optimum solution} & \textbf{Cost} & \textbf{G$_2$} & \textbf{G$_3$} & \textbf{G$_4$} \\
        \midrule
        \multirow{2}{*}{LLM-RBDO} & \multirow{2}{*}{Kriging} & [0.52, 1.35, 0.5, 1.38, 0.73,  & \multirow{2}{*}{25.59} & \multirow{2}{*}{0.989} & \multirow{2}{*}{0.9978} & \multirow{2}{*}{0.9545} \\
        & & 1.23, 0.58, 0.31, 0.26] & & & & \\
        \multirow{2}{*}{LLM-RBDO} & \multirow{2}{*}{True constraint} & [0.54, 1.34, 0.5, 1.39, 0.92,  & \multirow{2}{*}{25.81} & \multirow{2}{*}{0.999} & \multirow{2}{*}{0.998} & \multirow{2}{*}{0.91} \\
        & & 1.21, 0.57, 0.28, 0.23] & & & & \\
        \multirow{2}{*}{SEGA} & \multirow{2}{*}{Kriging} & [0.5, 1.39, 0.5, 1.31, 0.773,  & \multirow{2}{*}{25.21} & \multirow{2}{*}{0.8987} & \multirow{2}{*}{0.9697} & \multirow{2}{*}{0.9183} \\
        & & 1.44, 0.5, 0.224, 0.34] & & & & \\
        \multirow{2}{*}{SEGA} & \multirow{2}{*}{True constraint} & [0.5, 1.39, 0.5, 1.31, 0.771,  & \multirow{2}{*}{25.21} & \multirow{2}{*}{0.8987} & \multirow{2}{*}{0.997} & \multirow{2}{*}{0.9274} \\
        & & 1.44, 0.5, 0.228, 0.34] & & & & \\
        \multirow{2}{*}{FORM} & & [0.5, 1.31, 0.5, 1.24, 0.64,  & \multirow{2}{*}{24.14} & \multirow{2}{*}{1} & \multirow{2}{*}{0.999} & \multirow{2}{*}{0.9} \\
        & & 1.5, 0.5, 0.346, 0.192] & & & & \\
        \bottomrule
    \end{tabular}
\end{table}
\begin{table}[htbp]
    \centering
    \caption{Comparison of results for case study 2(b)}
    \label{table:4}
    \begin{tabular}{l c c c c c c}
        \toprule
        \textbf{RBDO method} & \textbf{Reliability estimation} & \textbf{Optimum solution} & \textbf{Cost} & \textbf{G$_8$} & \textbf{G$_9$} & \textbf{G$_{10}$} \\
        \midrule
        \multirow{2}{*}{LLM-RBDO} & \multirow{2}{*}{Kriging} & [0.52, 1.35, 0.5, 1.38, 0.73, & \multirow{2}{*}{25.59} & \multirow{2}{*}{0.9092} & \multirow{2}{*}{0.9993} & \multirow{2}{*}{0.9042} \\
        & & 1.23, 0.58, 0.31, 0.26] & & & & \\
        \multirow{2}{*}{LLM-RBDO} & \multirow{2}{*}{True constraint} & [0.54, 1.34, 0.5, 1.39, 0.92, & \multirow{2}{*}{25.81} & \multirow{2}{*}{0.9} & \multirow{2}{*}{0.999} & \multirow{2}{*}{0.998} \\
        & & 1.21, 0.57, 0.28, 0.23] & & & & \\
        \multirow{2}{*}{SEGA} & \multirow{2}{*}{Kriging} & [0.5, 1.39, 0.5, 1.31, 0.773, & \multirow{2}{*}{25.21} & \multirow{2}{*}{0.8937} & \multirow{2}{*}{0.9452} & \multirow{2}{*}{0.9638} \\
        & & 1.44, 0.5, 0.224, 0.34] & & & & \\
        \multirow{2}{*}{SEGA} & \multirow{2}{*}{True constraint} & [0.5, 1.39, 0.5, 1.31, 0.771, & \multirow{2}{*}{25.21} & \multirow{2}{*}{0.8915} & \multirow{2}{*}{1} & \multirow{2}{*}{0.9204} \\
        & & 1.44, 0.5, 0.228, 0.34] & & & & \\
        \multirow{2}{*}{FORM} & & [0.5, 1.31, 0.5, 1.24, 0.64, & \multirow{2}{*}{24.14} & \multirow{2}{*}{0.9} & \multirow{2}{*}{0.9892} & \multirow{2}{*}{0.9} \\
        & & 1.5, 0.5, 0.346, 0.192] & & & & \\
        \bottomrule
    \end{tabular}
\end{table}

\section{Conclusion}
This paper develops a novel reliability-based design optimization framework, termed LLM-RBDO, based on LLMs. The framework leverages the ICL capability of LLMs and integrates it with the iterative search mechanism of metaheuristic algorithms to enhance optimization efficiency and solution quality. To validate the effectiveness of this approach, LLM-RBDO is applied to two RBDO case studies and compared with the SEGA algorithm. The results indicate that LLM-RBDO achieves optimization performance comparable to SEGA and exhibits faster convergence in the first case. However, in the second case, LLM-RBDO shows some limitations due to the increased dimensionality of the optimization variables, suggesting that its applicability to high-dimensional optimization problems requires further investigation. Additionally, certain hyperparameters of LLM-RBDO have not been systematically fine-tuned, and further optimization of these parameters may improve its overall performance. Moreover, the sensitivity of this framework to LLMs with different parameter scales has not been fully explored. Future research could focus on experiments with various LLM sizes to assess their generalization capability and applicability. Overall, LLM-RBDO demonstrates significant potential and offers a novel approach to reliability-based design optimization. However, further improvements and refinements are necessary to enhance its applicability and robustness in complex engineering optimization problems.
\bibliographystyle{unsrt}  
\bibliography{paper/references}  %%% Remove comment to use the external .bib file (using bibtex).


% George Kour and Raid Saabne.
% \newblock Real-time segmentation of on-line handwritten arabic script.
% \newblock In {\em Frontiers in Handwriting Recognition (ICFHR), 2014 14th
%   International Conference on}, pages 417--422. IEEE, 2014.

% George Kour and Raid Saabne.
% \newblock Fast classification of handwritten on-line arabic characters.
% \newblock In {\em Soft Computing and Pattern Recognition (SoCPaR), 2014 6th
%   International Conference of}, pages 312--318. IEEE, 2014.

% \bibitem{hadash2018estimate}
% Guy Hadash, Einat Kermany, Boaz Carmeli, Ofer Lavi, George Kour, and Alon
% \newblock Estimate and replace: A novel approach to integrating deep neural
%   networks with existing applications.
% \newblock {\em arXiv preprint arXiv:1804.09028}, 2018.

% \end{thebibliography}

\end{document}